\documentclass[letterpaper, 10 pt, journal, twoside]{IEEEtran}
\usepackage[T1]{fontenc}

\usepackage[space, compress, sort]{cite}

\usepackage{amssymb} 
\usepackage{amsfonts}
\usepackage{amsthm}
\usepackage{wrapfig}
\usepackage[ruled,vlined]{algorithm2e}
\usepackage{mathtools}
\usepackage{tabularx}
\usepackage{graphicx}
\usepackage{enumerate}
\usepackage{float}
\usepackage{url}
\usepackage{verbatim}
\usepackage{booktabs} 
\usepackage{multirow}
\usepackage[linkcolor=black,citecolor=black,urlcolor=black,colorlinks=true]{hyperref}
\usepackage{bm}
\DeclareMathOperator*{\argmin}{arg\,min}

\usepackage{algpseudocode}

\SetKwRepeat{Do}{do}{while}
\newcommand{\tabincell}[2]{\begin{tabular}{@{}#1@{}}#2\end{tabular}}

\usepackage{subfigure}
\usepackage[normalem]{ulem}
\usepackage{xcolor} 
\newcommand{\delete}[1]{{\bgroup\markoverwith{\textcolor{red}{\rule[0.5ex]{2pt}{0.4pt}}}\ULon{#1}}}

\begin{document}

\title{Towards Optimizing a Convex Cover of Collision-Free Space for Trajectory Generation}

\author{Yuwei Wu, Igor Spasojevic, Pratik Chaudhari, Vijay Kumar

\thanks{
This work was supported by TILOS funded by NSF Grant CCR-2112665, USDA/NIFA Grant 2022-67021-36856, IoT4Ag ERC funded by NSF Grant EEC-1941529, and ONR Grant N00014-20-S-B001. 
}
\thanks{The authors are with the GRASP
Laboratory, University of Pennsylvania, Philadelphia, PA 19104 USA
{\tt\small\{yuweiwu, igorspas, pratikac, kumar\}@seas.upenn.edu}.}
}

\maketitle

\maketitle              

\begin{abstract}
We propose an online iterative algorithm to optimize a convex cover to under-approximate the free space for autonomous navigation to delineate Safe Flight Corridors (SFC).
The convex cover consists of a set of polytopes such that the union of the polytopes represents obstacle-free space, allowing us to find trajectories for robots that lie within the convex cover.
In order to find the SFC that facilitates trajectory optimization, we iteratively find overlapping polytopes of maximum volumes that include specified waypoints initialized by a geometric or kinematic planner.
Constraints at waypoints appear in two alternating stages of a joint optimization problem, which is solved by a novel heuristic-based iterative algorithm with partially distributed variables.
We validate the effectiveness of our proposed algorithm using a range of parameterized environments and show its applications for two-stage motion planning.
\end{abstract}

\section{Introduction}
\IEEEPARstart{A}{ll} motion and path planning algorithms require the representation of the environment for collision avoidance.
Different algorithms choose different representations. 
Exact methods map obstacles from the workspace into the configuration space, this gives guarantees of completeness and optimality but may be computationally intractable~\cite{Lav06}. 
Sampling-based methods do not need to represent the environment explicitly but they only offer probabilistic guarantees and may still require an excessively large number of samples in cluttered high-dimensional spaces~\cite{karaman2011}. 
A middle ground has become popular over the past few years, including convex approximation of the collision-free space. 
This is computationally attractive because it allows us to convert motion planning into an efficient optimization problem.
This problem is also tractable because collision-free space can be described by a set of linear constraints corresponding to each of the polytopes in the cover.

\begin{figure}[!ht]
      \centering
      \includegraphics[width=1\columnwidth]{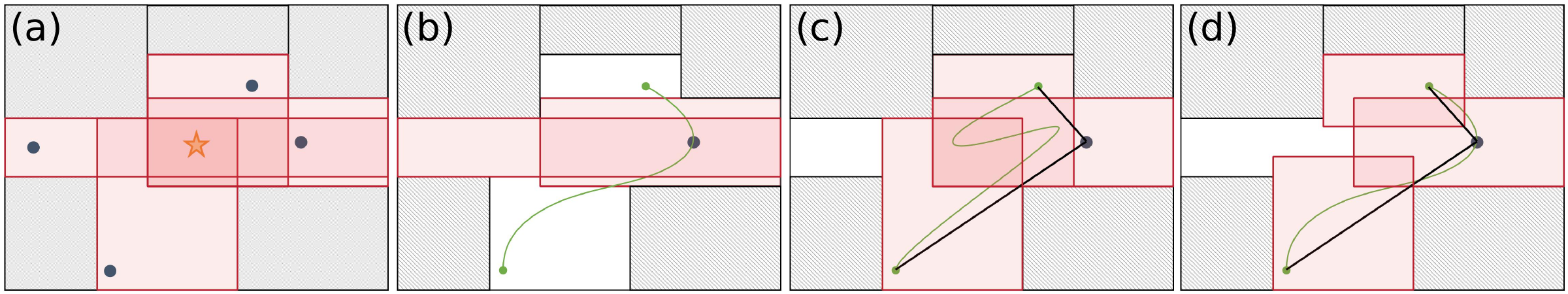}
      \caption{The influence of convex cover on trajectory generation.
      (a) Seed points (blue) determine the volume of the largest polytope (red) within a map (obstacles in grey). The largest-volume polytope can be identified by selecting the seed point at the orange star.
      (b) Polytopes may not always cover a large part of the trajectory, even if they have a large volume.
      (c, d) Given an initial path (black), the quality of the final trajectory (green) depends strongly upon which convex cover was used to optimize it.
      }
      \vspace{-0.2cm}
      \label{fig: fig1}
\end{figure}

\begin{figure}[!ht]
      \centering
      \includegraphics[width=0.9 \columnwidth]{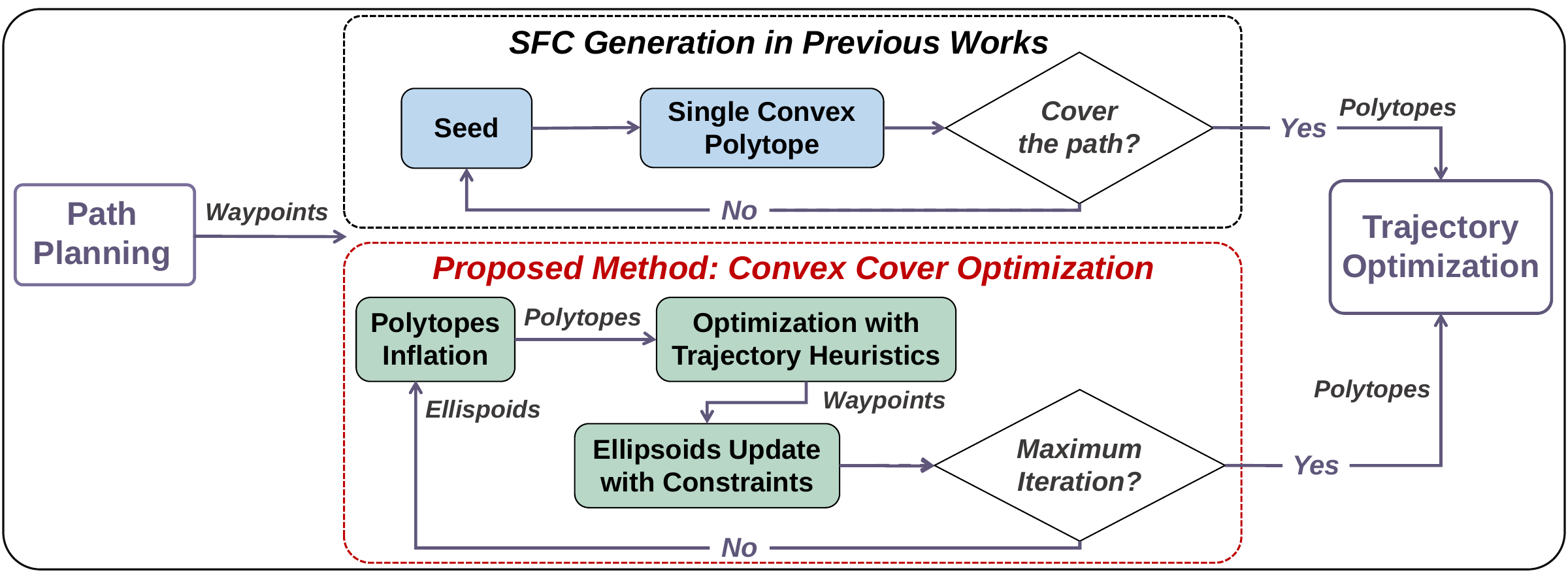}
      \vspace{-0.2cm}
      \caption{SFC generation pipeline of previous works and our method. Given waypoints from the path planning, the SFC generator selects seed points or lines from waypoints and outputs overlapping polytopes.
      The proposed method novelly formulates this process as a joint optimization problem to generate a better back-end trajectory downstream.}
      \label{fig: fig2}
\end{figure}

One of the key issues in this approach pertains to computing a good approximation of the environment. 
One could find a {\it convex cover}~\cite{1570649} by a collection of convex sets such that the union of the convex sets can cover a subset of the collision-free region or include a given guess for trajectories. 
Solving the subproblem of extracting a {\it single convex region} from a seed point or convex body is usually involved in generating the convex cover~\cite{Deits14computinglarge, 7784290, 10.1007/978-3-030-44051-0_3, zhong2020generating, wang2024fast}.
Considering the example in Fig.~\ref{fig: fig1}(a), an ideal single convex region should capture most of the free space. 
There should also be enough overlaps between different polytopes (ellipsoids are another popular choice) in the convex cover, at least successive polytopes as a “Safe Flight Corridor” (SFC).
The properties of the SFC will influence the ultimate geometric feasible region for optimization. 
Many existing approaches generate an SFC by selecting seed points around the initial path~\cite{7839930, 9531427, 9982032}.
However, such covers are not ideal in practice because the selection of SFC is greedy and does not account for trajectory planning.
It is difficult to determine how the geometric shapes and volumes of the polytopes interplay with the dynamical feasibility of the trajectories computed by the lower-level optimization procedure. 
Fig.~\ref{fig: fig1}(b)-(d) shows some examples of potential factors like polytope volumes and overlapping areas that can potentially change the final trajectory.
As a result, the existing pipelines (illustrated at the top part of Fig.~\ref{fig: fig2}) cannot give any guarantees on the quality of the SFC or the trajectory.
This suggests that coupled program: front-end path planning, free space approximation using a convex cover, and trajectory generation within this cover, cannot be solved independently.

To address the problem of SFC generation, we introduce a novel algorithm that optimizes the convex cover of the environment to aid in trajectory planning for quadrotors (and other mobile robots with dynamic constraints).
It updates the polytopes that form the convex cover and maximizes the volumes of ellipsoids inscribed within these polytopes given the waypoints updated by using heuristic functions to approximate trajectory cost.
These include two alternating stages of a joint optimization problem, which is solved by a heuristic iterative approach with partially distributed variables, inspired by the Alternating Direction Method of Multipliers (ADMM)~\cite{MAL-016} with partially distributed variables.
The pipeline is shown in Fig.~\ref{fig: fig2} in comparison to previous works.
This approach is validated using a large variety of synthetic environments to demonstrate its flexibility and robustness.

\textbf{Contributions:} (1) We formulate an SFC generation problem to optimize a convex cover of a subset of the collision-free space. This subset is approximated by the boundaries of trajectories that belong to the same homotopy class as a given path within the collision-free space for trajectory optimization.
(2) We design a novel iterative algorithm based on heuristics with partially distributed variables to maximize the efficiency of the computation.
(3) We extensively evaluate our method, benchmark it in environments with varying complexities, and present the first thorough analysis of the influence of the front-end plan path and SFC on the trajectory cost.

\vspace{-0.1cm}
\section{Related works}

In real-world representations of geometric environments, obstacles or objects can be decomposed into a series of convex shapes to assist obstacle avoidance strategies.
There are traditional methods to solve exact and approximate convex region~\cite{10.1145/800076.802459, 10.1145/1236246.1236265, wei2022approximate} or neural networks to learn the convex primitives \cite{9157370}.
However, decomposing obstacles would result in a large number of small pieces that can lead to significant computational costs.
Instead, extracting the convex representation of nonconvex free space is more efficient for motion planning problem formulation~\cite{7487283, marcucci2024fast, 9982032, 9812306, 9981518, werner2024approximating, gao2020teach}. 
Collision-free corridors or graphs can be generated using boxes\cite{7487283, marcucci2024fast}, sphere or ellipsoids\cite{9981518, 9812306} and polytopes\cite{9982032, werner2024approximating} within environments represented by point clouds or obstacle maps.

To extract convex regions from nonconvex environments, \cite{Deits14computinglarge} proposed an algorithm of Iterative Regional Inflation by Semidefinite Programming (IRIS) to generate maximum volume polytopes around convex obstacles given a seed point. 
The iterative optimization process involved sequentially quadratic programming to find separating hyperplanes toward obstacles, followed by semidefinite programming to determine the maximum-volume ellipsoid (inner \textit{Löwner-John Ellipsoid}~\cite{John2014}) within this polytope.
However, considering the expensive computation of this approach, most other works focused on improving the efficiency of the process, such as generating polytopes on line segments~\cite{7839930}, or convex bodies~\cite{wang2024fast}. 
With given voxel maps as inputs,~\cite{gao2020teach} proposed parallel convex inflation of axis-aligned cubes on axis directions by checking the voxel points incrementally, which can achieve resolution near real-time.
Moreover, in~\cite{7998590}, the authors introduced a stereographic projection technique to project obstacle points or point clouds and convert them into a convex hull.
The generative region is over-conservative for application, which is improved in~\cite{zhong2020generating} using sphere flipping to generate star-convex shapes and further refined into convex.
The work in \cite{wang2024fast} improves the polytope inflation and maximum volume inscribed ellipsoid processes in IRIS, providing a more efficient solution through a linear-complexity algorithm.
Except for inflating a single convex region within naturally nonconvex environments, determining a convex cover of local or global space remains challenging.
Several offline methods face challenges in selecting appropriate seed points to effectively generate a convex cover of the environment~\cite{Deits14computinglarge, 7138978}.
To improve the sampling efficiency, the work~\cite{10.1007/978-3-030-44051-0_3} leveraged the topology property of the objects and employed skeleton structures to position convex primitives. 
Furthermore, \cite{werner2024approximating} iteratively conducted sampling to update clique covers of visibility graphs and inflate polytopes. 
With the convex free-space maps, the trajectory generation can be addressed by mix integer programming~\cite{7138978, doi:10.1137/22M1523790}.

For real-world applications to trajectory optimization, computational efficiency is crucial and is typically achieved by decoupling the problem into subproblems or by narrowing the solution space. 
A two-stage planning framework, involving front-end path planning with geometric constraints followed by back-end trajectory optimization within the same homotopy class, has been extensively discussed in~\cite{10610207} and applied to SFC-decomposed environments.
The authors in \cite{7839930} proposed a jump point search (JPS) to find a geometric path composed of line segments and initialize ellipsoids on lines with one iteration polytope inflation.    
For general polytopes generation using seed points or lines, different paths from front-end pathfinding can be applied by iteratively checking the farthest point along the path within the polytope as the next sample points~\cite{9531427, 9982032}.
However, the above methods either fix line segments or employ a greedy approach to find the next waypoint as seeds, potentially resulting in suboptimal trajectories, particularly in narrow hallways.

Unlike previous methods, we introduce a novel approach that optimizes the selection of seeds in polytope generation for better trajectory optimization.
Our approach uniquely focuses on decomposing collision-free space to SFC, addressing critical factors such as convex approximation accuracy, trajectory segment division, and connectivity of convex regions between waypoints.
This integrated approach ensures that the trajectory optimization process is both optimized and computationally efficient, setting our method apart from existing techniques.

\vspace{-0.12cm}
\section{Problem Statement}

\subsection{Trajectory Optimization Formulation}

We consider the problem of finding an optimal energy minimization trajectory for a $s-$order integrator system, represented by states $x(t)$ and control inputs $u(t)$, with traversal time $T$ that passes through collision-free space $\mathcal{F}_{free}$.
Given a start state $x_0 \in \mathcal{F}_{free}$ and a feasible goal region $\mathcal{F}_{goal} \subseteq \mathcal{F}_{free}$, the trajectory optimization problem is formulated as 
\begin{subequations}
\begin{align}
\label{prob: opt_problem1}
& \min_{ x(t), T} \ \int_{0}^{T} \| u(t) \|^2 dt  + w_t \gamma (T) ,  \\
&  {\rm s.t.}  \ \ x^{(s)}(t)  = u(t),  \  \forall t \in [0, T], \label{eq: dynamics} \\ 
&  \qquad    x(0) = x_0,  \ \ x(T) \in  \mathcal{F}_{goal} , \label{eq: boundary condition}\\
&  \qquad  x(t) \in \mathcal{F}_{free}	, \  \forall t \in [0, T] \label{eq: free_space}.
\end{align}
\end{subequations}
The objective function minimizes a positive combination of the control effort and the total traversal time.
$\gamma(\cdot)$ is a time regularization function with weight $w_t $, that can encode several different requirements such as minimizing the total time, reaching the goal region at a target time, etc. 
The constraint (\ref{eq: dynamics}) represents robot dynamics, and $x^{(s)}$ denotes the $s$-th derivative of $x$.
The constraints (\ref{eq: boundary condition}) refer to start and end conditions.
This constrained optimization problem is nontrivial to solve due to the non-convexity of collision-free configuration space.  
\subsection{Convex Cover Generation}

We consider n-dimensional geometric space represented as $\mathcal{X}= \mathcal{X}_{free} \cup \mathcal{X}_{obs} \subseteq \mathbb{R}^n $, where $n \in \{2,3\}$. 
$\mathcal{X}$ consists of free space $\mathcal{X}_{free}$ and obstacles $\mathcal{X}_{obs}$, where $\mathcal{X}_{obs}$ is defined as a union of convex bodies $\mathcal{O} = \{\mathcal{O}_i\}_{i=1}^{S} $.
Several different representations exist for the convex bodies themselves. 
Typically they are either convex polytopes (bounded intersections of closed half-spaces) or spheres which are commonly used while manipulating raw point-cloud observations of the environment. 

In general, for navigation scenarios where there is enough free space in the environment, a conservative yet efficient strategy for preventing collisions involves uniformly inflating the grid map or obstacles and simplifying the robot to a point mass. 
Such pre-processing allows for trajectory planning directly within the geometric space to generate a collision-free geometrical path.
We consider an initial geometric path represented as a concatenation of straight line segments, whose endpoints form the sequence of waypoints $p = (p_0, \cdots, p_M)$,
\begin{equation}
\begin{aligned}
\Gamma(p) =\{x \in \mathbb{R}^n | \ x = \theta p_{i-1} + (1-\theta)p_{i}, \\
\forall \theta \in [0, 1], \
\forall i = {1, \cdots, M}    \} .
\end{aligned}
\end{equation}
With such a collision-free path, for example, output by a path planning algorithm, the convex cover problem can be formulated as finding a sequence of overlapping convex sets in collision-free space that covers the given path.
To obtain a parameterized family of convex sets suitable for optimization, we begin by defining a collision-free convex polytope using $\mathcal{H}$-representation $\mathcal{P}(A, b) = \{ x \in \mathbb{R}^n | Ax \leq b\} \subseteq  \mathcal{X}_{free}$, with $A \in \mathbb{R}^{v \times n}$ and $b \in \mathbb{R}^n$. 
We employ a sequence of overlapping polytopes for covering a subset of the collision-free space that contains the class of paths homotopic to the original path.
Hence, the convex cover of a given path is to generate an SFC such that
\begin{subequations}
\label{eq: convex_cover}
\begin{align}
\min_{\mathcal{P}}   & \ \ \sum_{i = 1}^M f(\mathcal{P}_i), \  \qquad  \quad  \label{fun: cost1} \\
 {\rm s.t.} \ \
   & \Gamma \in \bigcup_{i = 1}^{M} \mathcal{P}_i, \label{const: path1}\\
   & \mathcal{P}_i \cap \mathcal{P}_{i+1} \neq \varnothing, \ \  \forall 1 \leq i \leq M-1. \ \ \label{const: path2}
    \end{align}
\end{subequations}
We represent sequential polytopes $\mathcal{P} = (\mathcal{P}_1, \cdots, \mathcal{P}_M)$, with the $i$-th polytope $ \mathcal{P}_i$ parameterized by $A_{i} = [a_{i, 1}^T, \cdots , a_{i, v_i}^T]^T $ and $ b_{i} = [b_{i, 1}, \cdots , b_{i, v_i}]^T$.
The function $f(\cdot)$ defines an objective function to find the \textit{best} SFC.
Examples of possible criteria encoded by $f$ include the volume of polytopes $f(\mathcal{P}_i) =  - vol(\mathcal{P}_i)$, or the volume of the inner Löwner-John Ellipsoid of each polytope.
Nevertheless, while previous classes of criteria are intuitively appealing, it is not obvious how they lead toward synthesizing better trajectories. 
In the rest of the paper, we will discuss the factors and properties of SFC that are essential for effective trajectory generation.

\subsection{Trajectory Optimization with SFC}

With different representations of the dynamics and trajectories, we can formulate different optimization problems while minimizing control efforts. 
We will discuss one type of back-end optimization method that is applied to provide a heuristic cost function in our algorithm and thus can be extended to any back-end planner. 
If a trajectory is represented in flat-output space~\cite{5980409}, we can further characterize it by $2s-1$ degree $n$-dimensional piecewise polynomials 
\begin{equation}
\begin{aligned}
\sigma(t) = \{ \sigma_i(t) \ | \ t \in [\sum_{k = 1}^{i-1} \Delta t_{k}, \sum_{k = 1}^{i} \Delta t_{k}], \\
\quad \ \tau = [ \Delta t_1, \cdots,  \Delta t_M]^T \}.
\end{aligned}
\end{equation}
The $ \tau \in \mathbb{R}^M_{>0}$ represents a time interval vector with a total traversal time $ \mathbf{1}^T \tau  = T$, where $\mathbf{1}$ is a one-vector of size $M$. 
The time-optimal minimum control trajectory optimization with approximate convex cover can be formulated as a parametric nonlinear programming problem,
\begin{subequations}
    \label{prob: two}
    \begin{align}
	\min_{\sigma, \tau, A, b} & J(\sigma, \tau) = \int_{0}^{ \mathbf{1}^T \tau} \| \sigma^{(s)}(t) \|_2^2 dt + w_t  \gamma (\tau) , \label{eq:1}\\
	\ {\rm s.t.} \ \ 		
	&\ \  \sigma_1^{[s-1]}(0) = \bar{p}_0, \ \sigma_M^{[s-1]}( \Delta t_{M}) = \bar{p}_f,  \label{eq:2}\\
 	&\ \ \sigma_i^{[s-1]}(\Delta t_i) = \bar{p}_i, \ \forall i \in {1, \cdots , M},  \\
	&\  \ A_{i}\sigma_i(t) \leq b_{i} ,\ \forall i \in {1, \cdots , M}, \ \forall  t \in [0, \Delta t_i] , \label{ineq: corridor}  
    \end{align}
\end{subequations}
where $\bar{p}_0,   \bar{p}_i ,  \bar{p}_f \in \mathbb{R}^ {n \times s} $ are initial, intermediate and end derivative vectors, $\sigma_{\cdot}^{[s-1]} = (\sigma_{\cdot}(t), \cdots,  \sigma_{\cdot}(t)^{(s-1)})$.
As discussed in~\cite{wang2021generating}, the piecewise polynomial trajectory can be parameterized as \( \sigma(t) = \sigma(p, \tau)(t) \), where \( p = (\bar{p}_0, \cdots, \bar{p}_i, \cdots, \bar{p}_f) \). For brevity, we will use \( p \) and \( \tau \) to represent the parameters that determine the polynomial trajectory.
Trajectory optimization with piece-wise polynomial reduces the complexity of dynamics and scaling issues of discretizations, whereas introducing nonlinearity in objectives, and constraints coupled with coefficients, times, and SFC variables. 
Previous research in piece-wise trajectory with corridors mainly focuses on time allocation~\cite{5980409, 9147300, 10412114}, while the optimality and feasibility of convex approximation of environments are rarely discussed. 

\section{Proposed Algorithms}

In a single convex decomposition problem, an ellipsoid is typically coupled with the polytope during optimization~\cite{Deits14computinglarge}.
This is partly because the volume of the polytope is highly correlated with the volume of its inscribed ellipsoid, and optimizing the volume of the latter offers computational advantages.
To find the trajectory-optimal SFC for covering paths in the same homotopy equivalence class, 
we introduce a sequence of intersecting ellipsoids to encode a set of valid deformations of the current guess of the optimal path. 
For computational efficiency, we use the Cholesky representation of the ellipsoid and define it as $\mathcal{E}(L, d) = \{x \in \mathbb{R}^n | \  \| L^{-T} (x - d)  \|_{2} \leq 1 \} $, where $L \in \mathbb{R}^{n \times n }$ is lower triangular with strictly positive diagonal entries, and $d \in \mathbb{R}^n $ is the center of the ellipsoid.  
Our goal is to jointly find a geometric path together with a sequence of covering ellipsoids that simultaneously optimize the trajectory and the volume of the ellipsoids that certify it lies in free space. 
The problem is therefore formulated as
\begin{subequations}
\begin{align}
\label{eq: opt3}
\min_{  L_i, d_i, p_i, \forall i}   &  \  w_v   \tilde{f}( L_i)
 +  w_c  \tilde{J}(p, \tau), \  \qquad  \quad  \\
{\rm s.t.} \ \
& \| L_i^{-T} (p_i - d_i)  \|_{2} \leq 1, \  \forall i = 1, \cdots, M \label{eq:line1} \\
& \| L_i^{-T} (p_{i-1} - d_i)  \|_{2} \leq 1,  \ \forall i = 1, \cdots, M\label{eq:line2}
\\
& h_1(p, \tau, \mathcal{P}) \leq 0 , \label{eq:h1} \\
& h_2( \mathcal{E}(L, d),  \mathcal{P}) \leq 0 .\label{eq:h2}
\end{align}
\end{subequations}
The objectives are decoupled into two parts: the approximation of the polytope volume in Eq.~\eqref{fun: cost1} as the volume of inner ellipsoid $\tilde{f}(L)$, and an approximation of trajectory cost in Eq.~\eqref{eq:1} as $ \tilde{J}(p, \tau)$.
These terms are weighted by $w_v $ and $ w_c $ respectively.
The constraints~\eqref{eq:line1} and~\eqref{eq:line2} 
encode the requirement that every pair of consecutive waypoints lie in the corresponding ellipsoid. 
To rewrite the latter pair of constraints in a more streamlined way, we introduce the function $h_{0}(\mathcal{E}, p) = || L^{-T}(p - d)||_{2} - 1$.
The functions $h_1(\cdot), h_2(\cdot)$ in~\eqref{eq:line1} and~\eqref{eq:line2} represent separate conditions for ellipsoids and trajectory, which are further elaborated in section \ref{subsec: Waypoints Update} and section \ref{subsec: Maximum Volume Ellipsoids}.
We adopt the notation \( \mathcal{E} = (\mathcal{E}_1, \dots, \mathcal{E}_M) \), where each \( \mathcal{E}_i \) is parameterized by \( L_i \) and \( d_i \).
We reformulate the problem as follows:
\begin{subequations}
\begin{align}
\label{eq: opt1}
\min_{\mathcal{E}, p }   &    \  w_v   \sum_{i = 1}^{M}  \tilde{f}(\mathcal{E}_i) +  w_c  \tilde{J}(p, \tau),  \\
{\rm s.t.} \ \
&   g(h_0( \mathcal{E}_i , p_i)) = 0 , \ \forall i = 1, \cdots, M,  \\
&   g(h_0(\mathcal{E}_i, p_{i-1})) = 0, \ \forall i = 1, \cdots, M, \\
& h_1(p, \tau, \mathcal{P})  \leq 0 ,  \label{const: h_1}\\
& h_2(\mathcal{E}, \mathcal{P}) \leq 0, \ \label{const: h_2}\ 
\end{align}
\end{subequations}
where $g(\cdot)  = \max \{0, \cdot \}^2$ for coupled conditions of ellipsoid $\mathcal{E}$ and waypoint $p$ variables.
The independent constraints (\ref{const: h_1}) and (\ref{const: h_2}) are complex and may introduce additional variables such as trajectory time intervals and polytopes. 
Therefore, we incorporate the coupled constraints and formulate the scaled-form augmented Lagrangian as
\begin{equation}
\begin{aligned}
\mathcal{L}(\mathcal{E}, p, y) &=   w_v \sum_{i = 1}^{M}  \tilde{f}(\mathcal{E}_i) +  w_c   \tilde{J}(p, \tau) -  \frac{\rho}{2} \| y \|_2^2 \\
&+    \sum_{i = 1}^{M}   \frac{\rho}{2}  \| [g(h_0( \mathcal{E}_i , p_i))  ,g(h_0(\mathcal{E}_i, p_{i-1}))]^T +  y_i  \|_2^2 \\
\end{aligned}
\end{equation}
where $ y = (y_1, \cdots, y_{M}) $ are Lagrange multipliers, $\rho > 0 $ is a parameter. 
As our problem consists of decoupled subproblems, including convex region inflation, ellipsoid generation, and trajectory generation, multi-level decomposition \cite{mehrotra1992implementation, anandalingam1992hierarchical, 4118456, MAL-016} provides a suitable approach for efficiently solving the original problem.
To solve the problem, we propose an iterative algorithm, detailed in Alg. (\ref{alg: 1}), that leverages a scaled-form ADMM~\cite{MAL-016} in a partially distributed manner. 
We relax the constraints on independent variables of subproblems for efficient updates.
The input variables include an initial geometric path and obstacles, while the output is SFC represented by a sequence of polytopes.
The constraints $h(\cdot)$ would be satisfied during each sub-problem. 
Additionally, the cost function and constraints for ellipsoid generation are separable and can be updated in parallel.
\begin{algorithm}[!ht]
    \caption{Given an initial geometric path and obstacles, find an SFC to cover the path while avoiding all the obstacles.}
    \label{alg: 1}
    \KwIn{path $p^0$, obstacles $\mathcal{O}$}
    \KwOut{SFC $\mathcal{P}$}
    $\mathcal{E} \leftarrow    \rm{\textbf{InitEllipsoids}} $$(p^0)$ \\
    \Repeat{end condition}{
    $k \leftarrow  0$,  $\mathcal{P} \leftarrow \rm{\textbf{InflatePolytopes}}$($\mathcal{E}, \mathcal{O}$) \\
        \Repeat{$k = k_{max}$}{
         $p^{k+1}:= \argmin_{p} \{\mathcal{L}(\mathcal{E}^k, p, y^{k})  | \ h_1(p, \tau, \mathcal{P})  \leq 0     \} $ \\
          $\mathcal{E}_i^{k+1}:= \argmin_{\mathcal{E}_i} \{\mathcal{L}(\mathcal{E}_i, p^k, y_i^k) | \ h_2(\mathcal{E}_i, \mathcal{P}) \leq 0 \}, \ \forall i = 1, \cdots, M, $ \\
         $y_i^{k+1}:= y_i^{k} +  [ g(h_0(\mathcal{E}_i^{k+1} , p_i^{k+1})),  g(h_0(\mathcal{E}_i^{k+1} , p_{i-1}^{k+1}))]^T,  \ \forall i = 1, \cdots, M,  $\\
         $ k \leftarrow k +1$
        }
        $\mathcal{E} \leftarrow  \mathcal{E}^{k} $
    }
\end{algorithm}

\subsection{Initialization and Polytopes Generation}

Given an initial path $p^0$, we upsample waypoints by adding auxiliary intermediate waypoints between consecutive original waypoints to avoid creating excessively long ellipsoids that exceed a specified distance threshold $\alpha$.
We use the procedure \textbf{InitEllipsoids} that takes as input line segments of the initial path and outputs the ellipsoid centered at $d_i = (p_i + p_{i-1})/2$.
Let $\mu_i = \frac{p_{i} - p_{i-1}}{|| p_{i} - p_{i-1} ||{2}}$ be the unit vector along the displacement from waypoint $p_{i-1}$ to waypoint $p_{i}$. Then $L_{i}$ is chosen so that 
\begin{equation}
L_{i} L_{i}^T = \mu_{i} \mu_{i}^T \frac{|| p_{i} - p_{i-1} ||_2^2}{4} + (I - \mu_{i} \mu_{i}^T) \epsilon^2.
\end{equation}
We assume the initial path has  $\epsilon$-clearance from obstacles; this implies that the constructed ellipsoids are collision-free.
In \textbf{InflatePolytopes}, to find a maximum volume of polytopes in $\mathcal{X}_{free}$, we reformulate the optimization problem (\ref{eq: convex_cover}) as
\begin{subequations}
\begin{align}
\label{eq: opt2}
\min_{\mathcal{P}}   & \ \ \sum_{i = 1}^{M}  - vol(\mathcal{P}_i), \  \qquad  \quad  \\
 {\rm s.t.} \ \
   & \mathcal{E}_i \subset \mathcal{P}_i ,  \mathcal{P}_i \cap \mathcal{O} = \varnothing
   \quad \forall i = 1, \cdots, M. \ \
    \end{align}
\end{subequations}
We adopt a method similar to~\cite{Deits14computinglarge} to solve this problem by finding tangent separating planes from the ellipsoid.
To save computation, we define a local bounding box by adding a margin of $ \pm  l$ around the ellipsoid for each polytope generation. 
This margin ensures that only the \textit{local} obstacles within this expanded range are considered.
The overlap condition $\mathcal{P}_i \cap   \mathcal{P}_{i+1} \neq \varnothing$ is satisfied by construction.
Additionally, we store the obstacle points near each initialized ellipsoid for parallel updating of polytopes during iteration.

\subsection{Waypoints Update with Trajectory Heuristics}
\label{subsec: Waypoints Update}

The SFC serves as the environmental approximation for trajectory optimization, therefore the trajectory cost is the key factor during this phase. 
For back-end optimization, minimum control effort (energy), distance, and traversal time are three main factors. 
The optimal control trajectory $\sigma(t)$ is therefore encoded through its waypoints and time intervals $ \{p, \tau \}$, which formulates the problem
\begin{subequations}
\begin{align}
\label{eq: max_wp}
\min_{p, \tau }  \  &w_c   \tilde{J}(p, \tau ) +    \\
&\sum_{i = 1}^{M} \frac{\rho}{2}  \| [g(h_0( \mathcal{E}_i^k , p_i))  ,g(h_0(\mathcal{E}_i^k, p_{i-1}))]^T +  y_i  \|_2^2, \nonumber \\
{\rm s.t.} \ \
& h_1(p, \tau , \mathcal{P}) \leq 0.
\end{align}
\end{subequations}
Incorporating the energy cost introduces the dynamic model of the robot, which is complex to accurately derive and expensive to optimize simultaneously during SFC generation.
In the rest of the section, we will discuss the approximation of the objective $ \tilde{J}$ through minimum path length (denoted as $ \tilde{J}_1$) and minimum control effort ($ \tilde{J}_2$) for updating waypoints.

\subsubsection{Minimum Path Length}
Without the consideration of dynamic feasibility, distance serves as a common criterion for geometric pathfinding. 
We employ Euclidean distance between waypoints to find the shortest path inside the corridors and ellipsoids, as 
\begin{subequations}
\begin{align}
\label{eq: min_dist}
\min_{p }  \ &  w_c  \sum_{i = 1}^{M} \| p_i - p_{i-1}\|_2 +   \\  &\sum_{i = 1}^{M} \frac{\rho}{2}  \| [g(h_0(\mathcal{E}_i^k , p_i)) , g(h_0(\mathcal{E}_i^k , p_{i-1}))]^T +  y_i^k  \|_2^2,  \\
{\rm s.t.} \ \
& p_i \in \mathcal{P}_i \cap    \mathcal{P}_{i+1}, \ \forall i = 1, \cdots, M-1, \label{const: inside}
\end{align}
\end{subequations}
where constraint (\ref{const: inside}) ensures waypoints are generated within the overlapping region of every two consecutive polytopes.
\begin{figure}[!ht]
       \vspace{-0.15cm}
      \centering
       \includegraphics[width=0.9\columnwidth]{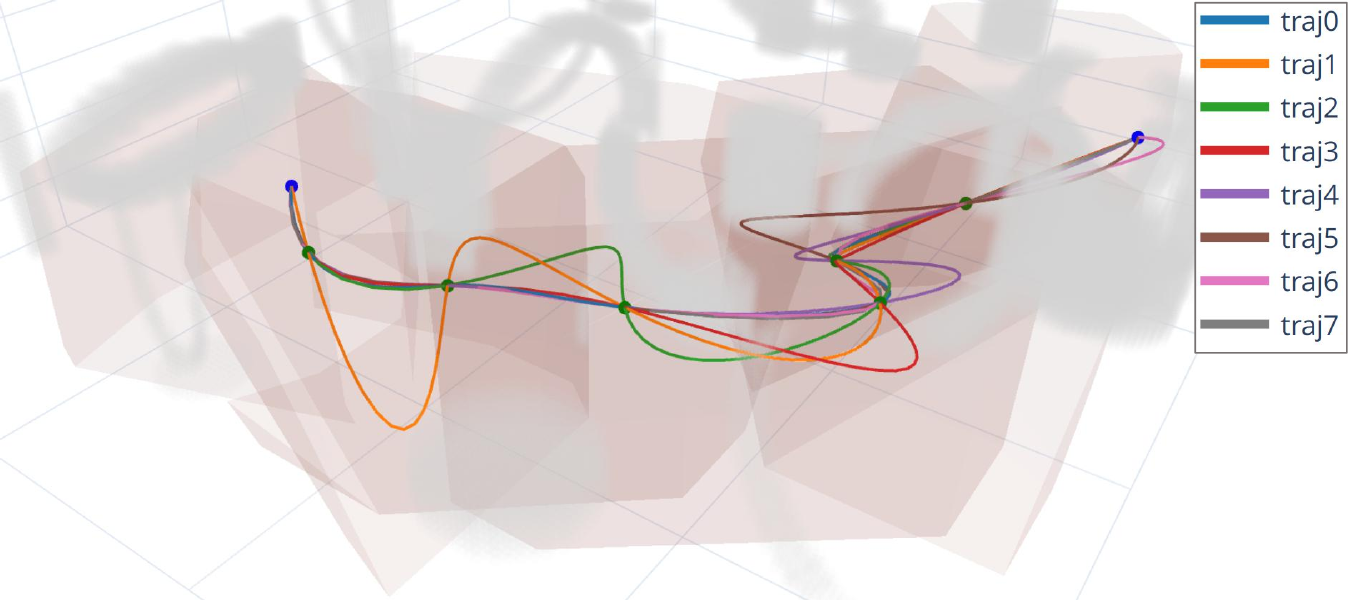}
       \vspace{-0.25cm}
      \caption{Polynomial trajectories can take on various shapes by changing time allocations.} 
      \label{fig: time_allocation}
\end{figure}

\subsubsection{Minimum Control Effort} 
The time allocation significantly influences the control efforts and the shape of the polynomials, shown in Fig.~\ref{fig: time_allocation}.
Including time allocation in optimization will also introduce nonlinear constraints. 
In addition, time allocation is also correlated to dynamic limits like maximum acceleration and velocities, which need to be specified. 
However, in rest-to-rest trajectory planning, one can always adjust a time factor to re-scale the time allocation to push the trajectory inside corridors.
In this case, we assume the rest-to-rest scenarios and solve trajectory optimization with geometric constraints as
\begin{subequations}
\begin{align}
\label{eq: min_con}
\min_{p }  \ &  w_c  \int_{0}^{ \mathbf{1}^T \tau} \| \sigma(p, \tau)^{(s)}(t) \|_2^2 dt + \\
&\sum_{i = 1}^{M} \frac{\rho}{2}  \| [g(h_0(\mathcal{E}_i^k , p_i)) , g(h_0(\mathcal{E}_i^k , p_{i-1}))]^T +  y_i^k  \|_2^2, \\
{\rm s.t.} \ \
& M \sigma(p, \tau) = e,   \label{const: bound}  \\
& A_i \sigma_i(p, \tau) \leq b_i,   \ \forall i = 1, \cdots, M \label{const: corridor},
\end{align}
\end{subequations}
where $\sigma_i$ is the $i$-th trajectory segment mapping from derivative vectors and time intervals, which has been discussed in \cite{richter2016polynomial}. 
The constraints (\ref{const: bound}) include boundary and continuity constraints for polynomials, and constraints (\ref{const: corridor}) enforces trajectory inside corridors. 
Practically, we can use a fixed time allocation with a time factor to provide sub-optimal heuristics.

\subsection{Maximum Volume Ellipsoids with Inner\&Outer Constraints}
\label{subsec: Maximum Volume Ellipsoids}

\begin{figure}[!ht]
      \centering   
      \includegraphics[width=0.9\columnwidth]{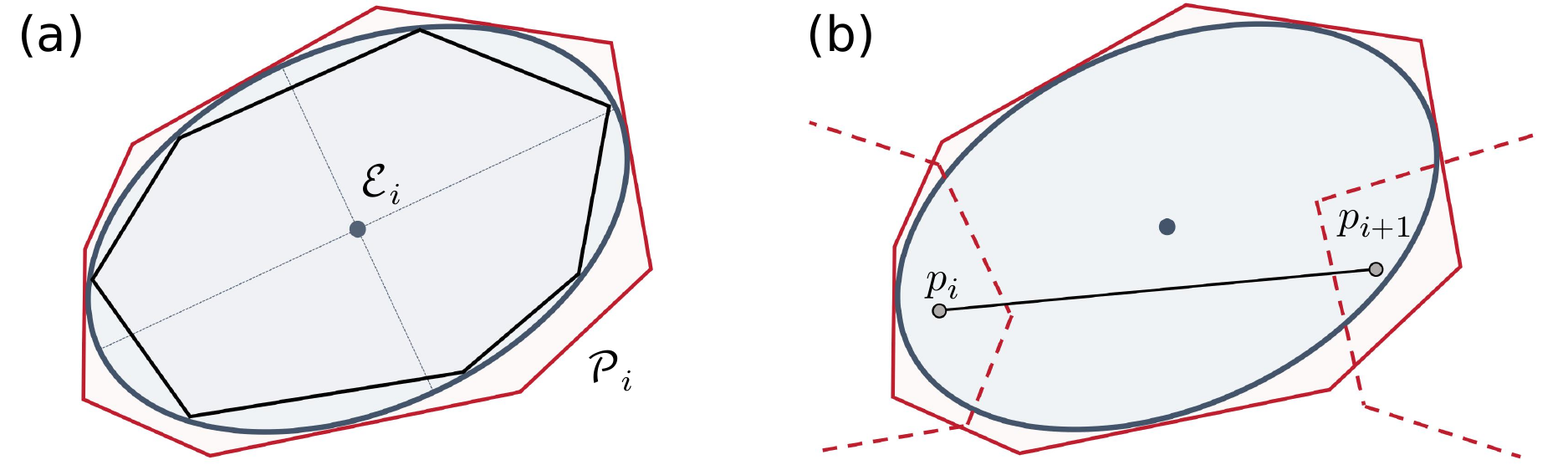}
      \caption{Demonstration of geometric constraints for an ellipsoid (blue). 
        (a) The general formulation of inner and outer Löwner-John ellipsoid.  (b) The Inner Löwner-John ellipsoid with outer line segment constraints.}
      \label{fig: all_ell}
\end{figure}

We introduce a cost function related to the volume of the convex region that can inflate to increase the potential solution space for trajectory optimization.
To maintain the connected constraints for every convex region, we have to ensure the overlapping conditions of polytopes or ellipsoids.
For each sub-problem, the goal is to find the maximal volume inscribed ellipsoid of a convex body (polytope) and the volume circumscribed ellipsoid outside a convex body (a line segment or multiple convex bodies).
As illustrated in Fig.~\ref{fig: all_ell}, an ellipsoid is constrained inside the polytope with different outer conditions. 
Given current waypoints $p^k$, we can formulate the sub-problem of finding the $i$-th ellipsoid $\mathcal{E}_i(L_{i}, d_i)$ by
\begin{subequations}
\begin{align}
\label{eq: max_ell}
\min_{L_{i}, d_i}  \ & - w_v \log \det  (L_{i}) +  \\
&\frac{\rho}{2}  \| [g(h_0(\mathcal{E}_i , p_i^k)), g(h_0(\mathcal{E}_i , p_{i-1}^k))]^T +  y_i^k  \|_2^2, \\
{\rm s.t.} \ \
&\| (a_{i, j})^{T} L_{i} \|_2 + (a_{i, j})^{T} d_i \leq b_{i, j}, \  \  j = 1, \cdots, v_i  \label{const: corridor2} \\
&L_{i} \ {\rm is \ lower \ triangular},  \ \ \
\end{align}
\end{subequations}
where the first term of objective, $\log \det (L_{i})$,
is proportional to the volume of the ellipsoid, and (\ref{const: corridor2}) represents corridor constraints.
We employ a distributed style to update $\mathcal{E}^{k+1}$.

\section{Results}

\subsection{Implementation Details}

We employ the environments provided in~\cite{10610207} which include both simulated and real-world point clouds for cluttered obstacle maps, maze maps, and real-world maps.
We follow their procedure which defines a quantity called the Environmental Complexity Signature (ECS) to generate simulated environments of varying complexities (obstacle structures, densities, etc).
All environments are stored as voxel maps.
We use the RRT*~\cite{karaman2011} algorithm from the OMPL library~\cite{sucan2012the-open-motion-planning-library} to get the initial geometric path for all methods.
We use the L-BFGS algorithm~\cite{liu1989limited} to minimize the augmented Lagrangian for all the constrained optimization problems in our approach and replace the rest hard constraints with penalties.
In practice, we run the inner iteration once and update the polytope through several outer iterations.   
Our approach can be deployed on our custom hardware platform~\cite{10160295} to provide an environmental representation for two-stage motion planning. 
\footnote{We will open-source at \url{https://github.com/KumarRobotics/kr_opt_sfc}.}

\subsection{Numerical Analysis}

We conduct several tests in a parameterized environment of size 20 $\times$ 20 $\times$ 5 m$^3$ with randomly chosen start and end points, at least 10 m apart.
All obstacles are inflated by the radius of the robot. 
We use total volumes of ellipsoids ($vol(\mathcal{E})$) and polytopes ($vol(\mathcal{P})$), and overlapping volumes of each neighboring pair of polytopes ($vol(\cap \mathcal{P})$) to showcase the performance of the proposed algorithm during several iterations. 
For evaluation among different test cases, the values were normalized by the maximum values observed during the iterations of a trial.

\begin{figure}[!ht]
    \centering
    \includegraphics[width=\linewidth]{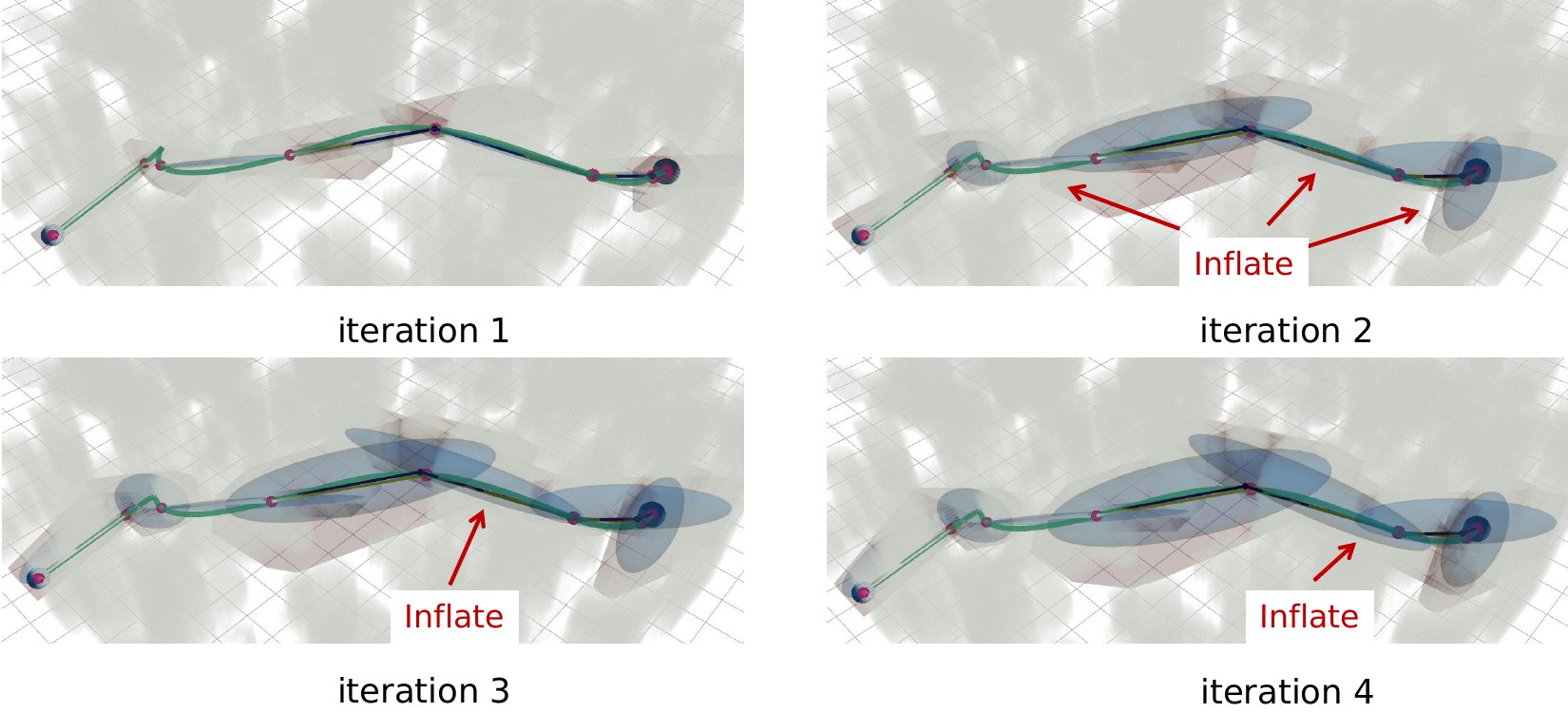}    
    \caption{Visualization of ellipsoids (blue) and polytopes (red) during 4 iterations. The black line segments represent the path, with waypoints in magenta. The green curve is the trajectory generated within polytopes.}
    \label{fig: sim}
\end{figure}
\subsubsection{Ellipsoids and Polytopes over Iterations.}
\begin{figure}[!ht]
    \centering
    \includegraphics[width=\linewidth]{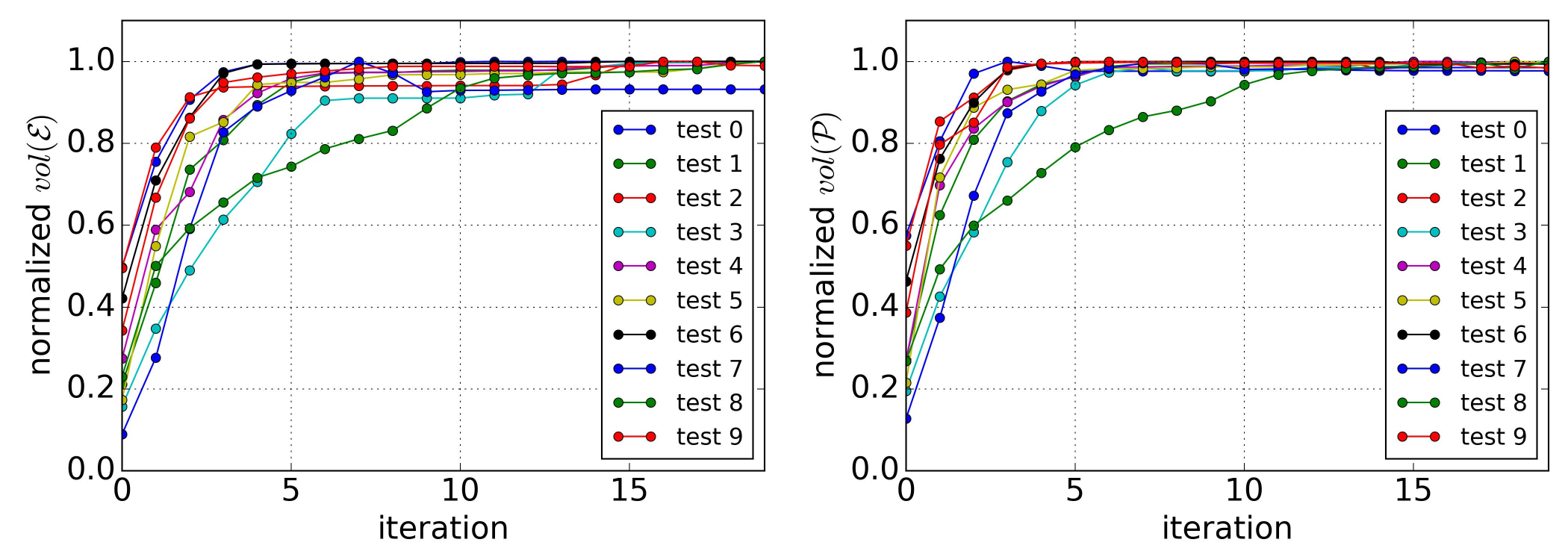}
    \caption{The variance in normalized volumes of ellipsoids and polytopes after 20 iterations of our proposed algorithm (across ten trials).}
    \label{fig: iter}
\end{figure}

To validate the effectiveness of our proposed method, we show one example using heuristics $\tilde{J}_1 $ and visualize the ellipsoids and polytopes of four iterations in Fig.~\ref{fig: sim}.
The center and orientation (or axes) impact the maximum volume an ellipsoid can inflate, and our proposed method is able to adjust it with iterations.
Fig.~\ref{fig: iter} illustrates the actual volumes of additional tests.
The normalized volumes of ellipsoids (left) and polytopes (right) consistently grow with each iteration and contain more than 90\% of the maximum volumes within 5-10 iterations in most cases.
In what follows, we will cap the number of iterations between 5 and 10; we empirically observed that this comes without sacrificing performance.
\begin{figure}[!ht]
    \centering
    \includegraphics[width=\linewidth]{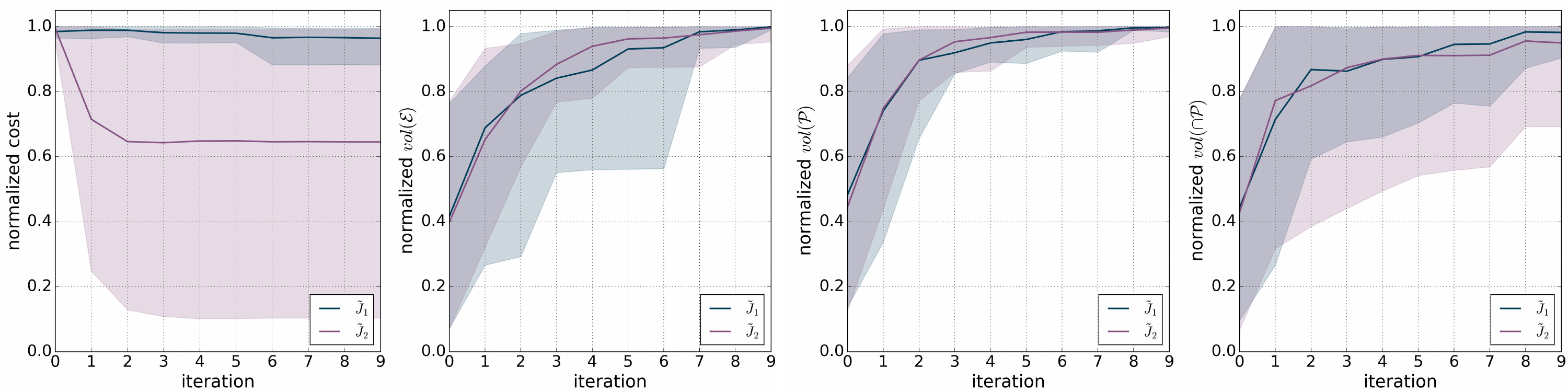}
    \caption{Comparison of performance by trajectory heuristics based on minimum path length ($ \tilde{J}_1$) and minimum control effort ($ \tilde{J}_2$) over 100 test cases.}
    \label{fig: iter_compare}
\end{figure}
\subsubsection{The Effects of Cost Heuristics.}
We compare two trajectory heuristics with minimum path length and minimum control effort (using jerk) for waypoint updates and show the dependence of normalized volumes of ellipsoids, polytopes, and overlapping areas of the convex cover with the number of iterations. 
We use a uniform time allocation proportional to the path length in minimum control effort to ensure that the waypoints are distributed uniformly as they move and that the cost term is updated consistently.
The performances using different trajectory heuristics costs are shown in Fig.~\ref{fig: iter_compare}.
The cost $ \tilde{J}_2$ performs slightly better than $ \tilde{J}_1$, regarding the coverage rate of ellipsoids and polytopes volumes.  
The cost value of distance has less margin to decrease as the initial path from RRT* has already minimized the distance to a large extent. 
\begin{figure}[!ht]
    \centering
    \includegraphics[width=0.97\linewidth]{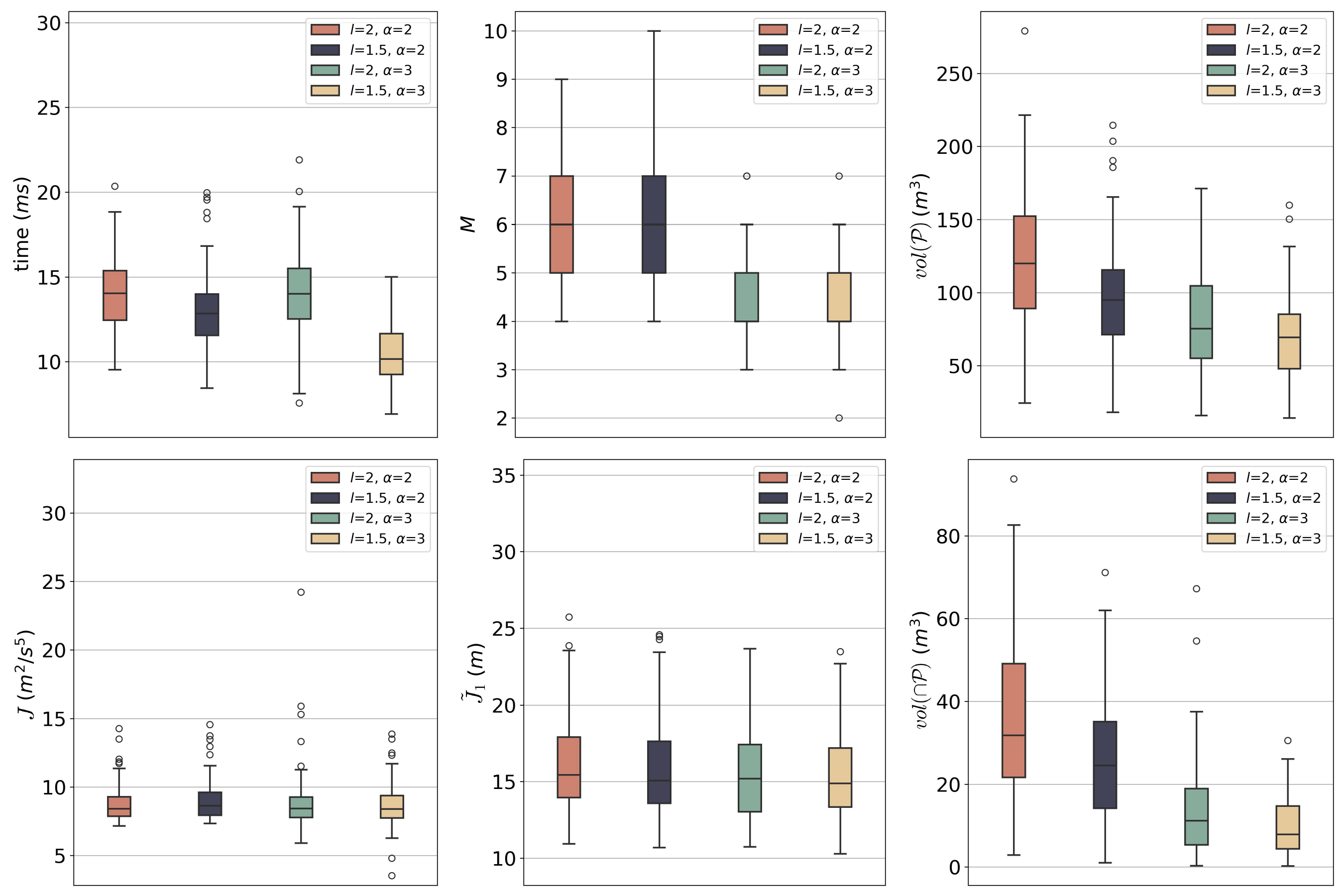}
    \caption{Performance by different settings with $l \in \{ 1.5, 2 \}$ and $\alpha \in \{2, 3\}$.}
    \label{fig: config}
\end{figure}

\subsubsection{Configurable Performance in Different Environments}
We leverage parameters such as local range $l$ and upsample threshold $\alpha$ to generate SFC with specified accuracy and efficiency.
Our analysis considers the following four setups with a local range of 1.5 m and 2.0 m and an upsample threshold of 2.0 m and 3.0 m.
The upsample threshold influences the number of segments, and the local range determines the number of obstacles considered for each polytope generation. 
These settings are evaluated in terms of computation time, trajectory cost ($J$), number of segments ($M$), path length ($\tilde{J}_1$), volumes of polytopes, and overlapping areas. 
Trajectory cost is evaluated by optimizing time intervals and waypoints in \cite{9765821}.
As shown in Fig.~\ref{fig: config}, the computation time can be reduced by decreasing the local range and increasing the upsample threshold.
The path length, polytope volumes, overlapping volumes, and trajectory cost also decrease accordingly. 
We conclude that within the same homotopic class, the optimal trajectory is indeed influenced by the convex cover.

\subsection{Comparison of Convex Cover}
\begin{figure}[!ht]
    \centering
    \includegraphics[width=\linewidth]{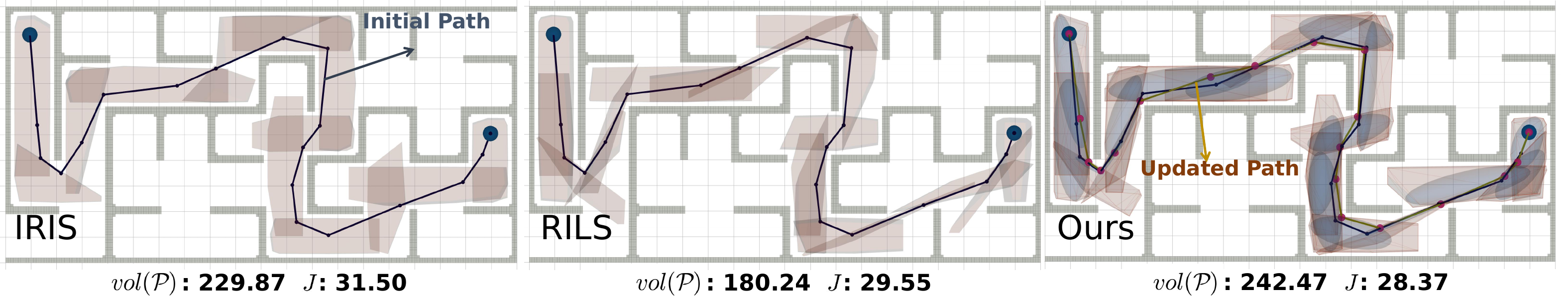}
    \caption{The illustration of SFC generated by three methods. More comparisons are available at \url{https://youtu.be/k7CI7-fgSXE}}
    \label{fig: benchmark3}
    
\end{figure}

We compare our proposed algorithm with two state-of-the-art SFC generation methods, IRIS\cite{Deits14computinglarge} and the method presented in \cite{7839930}, denoted as RILS.
IRIS was originally designed for graph cover, and we use its applications for a convex cover of a path with incremental point collision checking. 
The iteration number of IRIS is set as the major iteration in their paper, the same as our proposed method. 
To get inner Löwner-John ellipsoids in IRIS, we directly employ the Cholesky representation for optimization in~\cite{wang2024fast}. 
The local range for considering obstacles within the bounding box is set to $l = 2$ m to optimize computational efficiency.
We first evaluate the performance of three methods across three types of randomly sampled maps generated from~\cite{10610207}. 
We conduct 1,000 trials for each type of environment, with start and end points randomly selected, ensuring a minimum distance of 10 meters between them.
As shown in Table~\ref{tab: detailed_comparsion}, our method outperforms other benchmark methods for generating better trajectories, specifically in unstructured environments.
\begin{table}[!htbp]
    \centering
    \renewcommand\arraystretch{1.1}
    \setlength{\tabcolsep}{6pt}
    \caption{Comparison on Different Types of Maps. \label{tab: detailed_comparsion}}
\begin{tabular}{l|c|c|c|c|c}
\hline
\multirow{1}*{Map}  & Method &  \tabincell{c}{$J$ \\  ($m^2/s^5$)} & \tabincell{c}{ $vol(\mathcal{P})$\\ ($m^3$)} & Time ($ms$) & $\tilde{J}_1$  ($m$)  \\
\hline
\multirow{3}*{Obstacle}&  IRIS* & 35.39 & \textbf{107.82 }& 45.93 & 20.17 \\
&RILS   & 45.76 & 60.08 & 11.52  & 19.37 \\
&Ours  & \textbf{23.83} & 84.79 & 24.76 & 19.66  \\
\hline
\multirow{3}*{Maze} &  IRIS* & 33.82 & 118.38 & 33.85 & 20.31 \\
&RILS   & 33.48 & 83.93 & 6.06 & 18.81 \\
&Ours  & \textbf{33.02} & \textbf{121.79} & 30.46 & 19.56 \\
\hline
\multirow{3}*{Real}  & IRIS* & 20.53 & 144.86 & 17.68 & 14.04 \\
&RILS  & 28.28 & 158.45 & 3.11 & 13.80 \\
&Ours  & \textbf{16.45} & \textbf{178.72} & 19.32 & 14.04 \\
\hline
\multicolumn{4}{l}{*IRIS search range is cropped.} \\
\end{tabular} 
\end{table}
We visualize one trial as an instance, illustrated in Fig.~\ref{fig: benchmark3}.
Our method iteratively refines the path for better trajectory generation, particularly in complex environments such as narrow mazes where RILS show poor performance around corners.

\begin{figure}[!ht]
      \centering
    \includegraphics[width=\linewidth]{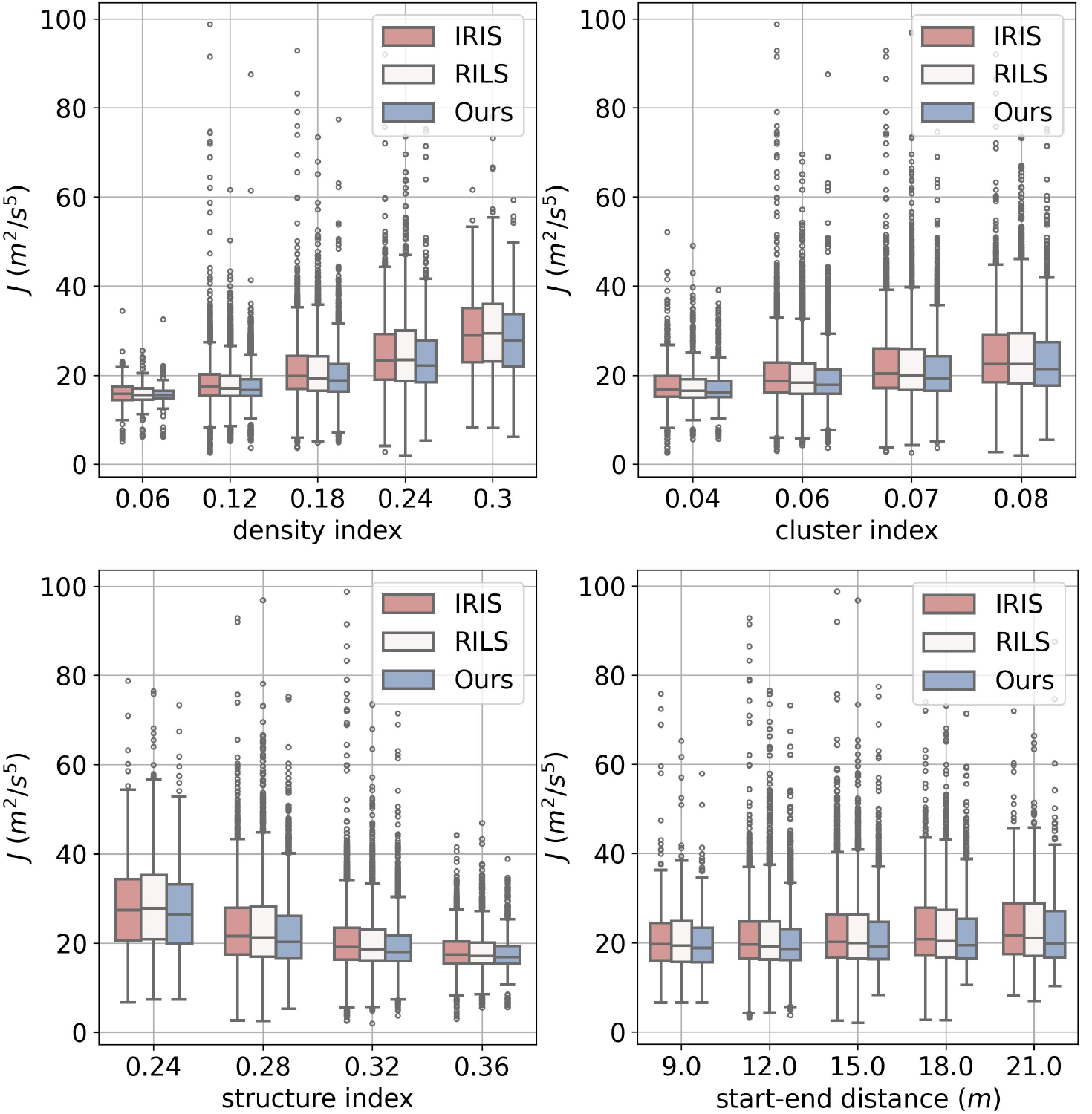}
    \caption{Benchmark comparison with different environmental parameters, including density index, cluster index, structure index (as defined in~\cite{10610207}) and start-to-end distance.}
    \label{fig: benchmark}
\end{figure}

It is clear that the performance of any algorithm is influenced by the complexity of both the environment and the task, which makes it difficult to draw conclusions based on average performance.
To further compare with benchmark methods, we conduct 10,000 additional tests in simulated environments characterized by the ECS (density index, cluster index, and structure index). 
The results are filtered to exclude cases where any method fails to find feasible corridors.
We categorize test cases based on the range of the density index to represent environment complexity and show the results in Fig.~\ref{fig: benchmark}.

Overall, our method achieves lower trajectory costs across various environments.
As density and cluster indices increase, environments become more complex, which leads to increased trajectory costs across all three methods. 
In contrast, more structured environments allow for trajectory generation with lower costs.
The comparison results categorized by start-to-end distance (the Euclidean distance between the start and end points), show that our method outperforms others as the distance increases.
More importantly, this is \textit{the first systematic analysis of multiple factors that influence the trajectory cost}: the waypoint and number of segments in the front-end plan, the volumes of the polytopes in the convex cover, and the volumes of consecutive polytopes.

\section{Conclusion and Future Work} 
\label{sec:conclusion}

This paper introduces a novel algorithm for optimizing the convex cover of collision-free space as SFC for trajectory optimization.
The convex cover is formulated as a joint optimization of ellipsoids, polytopes, and intermediate waypoints, which are updated iteratively to find optimal solutions.
The proposed method can be applied to two-stage motion planning as an intermediate process, or optimizing with trajectory generation jointly.
Our future work will address whole-body trajectory planning for $SE$(2) and $SE$(3).

\bibliography{references}
\end{document}